\documentclass[letterpaper]{article} % DO NOT CHANGE THIS
\usepackage{aaai20}  % DO NOT CHANGE THIS
\usepackage{times}  % DO NOT CHANGE THIS
\usepackage{helvet} % DO NOT CHANGE THIS
\usepackage{courier}  % DO NOT CHANGE THIS
\usepackage[hyphens]{url}  % DO NOT CHANGE THIS
\usepackage{graphicx} % DO NOT CHANGE THIS
\usepackage{graphicx}  % DO NOT CHANGE THIS
\usepackage[utf8]{inputenc}
\usepackage{bbold}
\usepackage{amssymb}
\usepackage{amsmath}
\usepackage{balance}
\usepackage{tikz}
\usepackage{subfig}
\usepackage{bbm}
\usepackage{enumitem}
\usepackage{appendix}

\DeclareMathOperator*{\argmax}{arg\,max}

\title{A Probabilistic Simulator of Spatial Demand for Product Allocation}

\author{Porter Jenkins \textsuperscript{\rm 1}, Hua Wei \textsuperscript{\rm 1}, J. Stockton Jenkins \textsuperscript{\rm 2}, Zhenhui Li \textsuperscript{\rm 1} \\ 
\textsuperscript{\rm 1} Penn State University \\
\textsuperscript{\rm 2} Brigham Young University \\%If you have multiple authors and multiple affiliations
}

\begin{document}
\maketitle

\begin{abstract}
Connecting consumers with relevant products is a very important problem in both online and offline commerce. In physical retail, product placement is an effective way to connect consumers with products. However, selecting product locations within a store can be a tedious process. Moreover, learning important spatial patterns in offline retail is challenging due to the scarcity of data and the high cost of exploration and experimentation in the physical world. To address these challenges, we propose a stochastic model of spatial demand in physical retail. We show that the proposed model is more predictive of demand than existing baselines. We also perform a preliminary study into different automation techniques and show that an optimal product allocation policy can be learned through Deep Q-Learning. 

\end{abstract}

\section{Introduction}

A key challenge for many physical retailers is choosing where to display their products. In many large stores, it can be difficult for consumers to find what they are looking for since a typical retailer may sell thousands of products. Additionally, consumers often purchase goods that they had not intended to buy beforehand, but are made on an impulse. Proper placement reduces search costs and maximizes "impulse" buys \cite{Badgaiyan}. For example, suppose a shopper visits a supermarket intending to purchase groceries. As the shopper checks out he sees a soft drink beverage placed near the cash register, and adds it to his cart. The shopper's decision to purchase the drink was in part a function of the environmental cues and placement of the product \cite{Mattila}. The main idea of this work is propose a strategy for automating the decision process of product placement in an optimal way.

\begin{figure}[t!]
\centering
\subfloat[Example floor plan]{\includegraphics[width=0.8\linewidth]{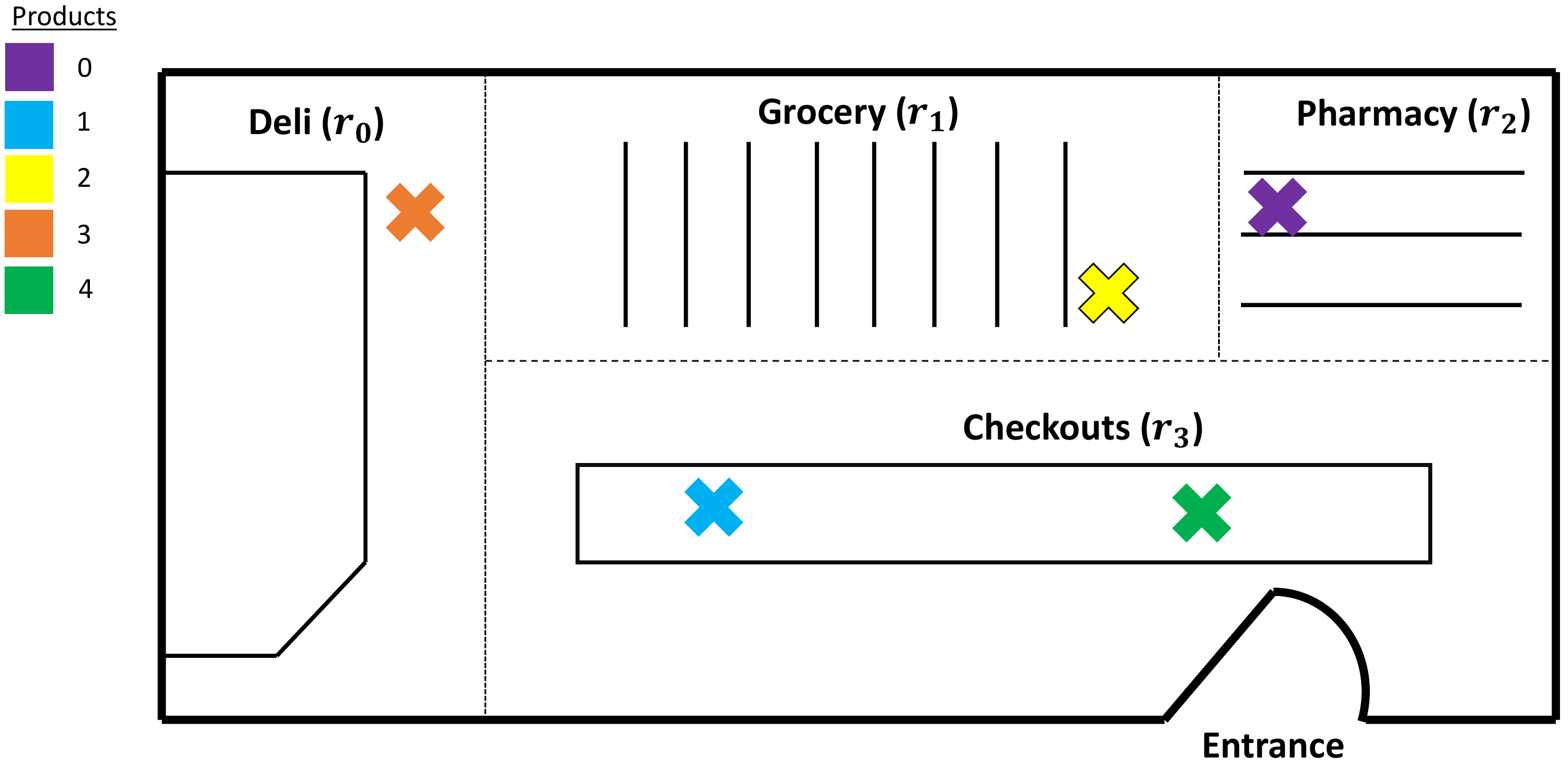}} \\
\vspace{-4mm}
\subfloat[State matrix]{\includegraphics[width=0.5\linewidth]{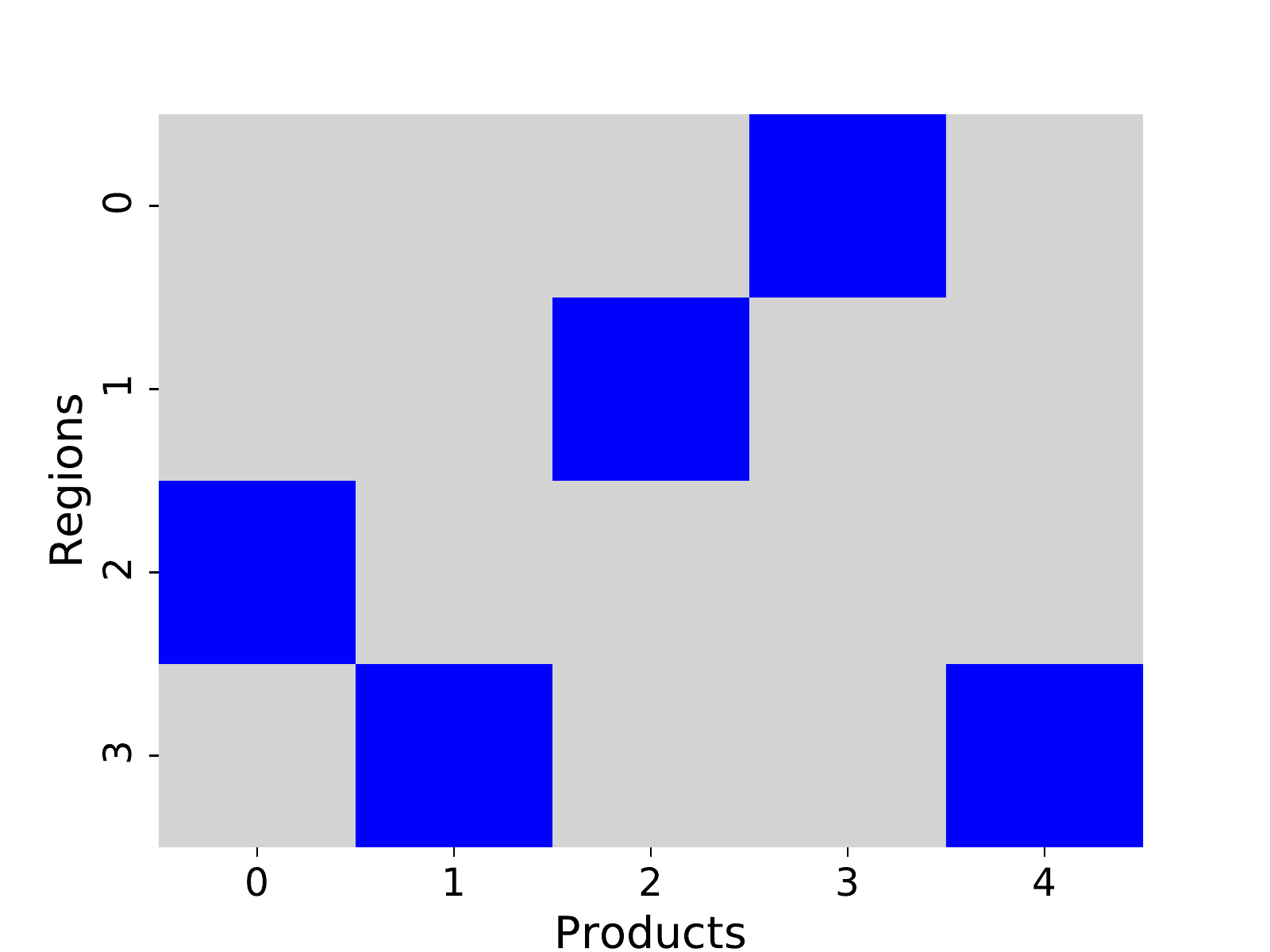}}
\subfloat[Revenue distribution]{\includegraphics[width=0.5\linewidth]{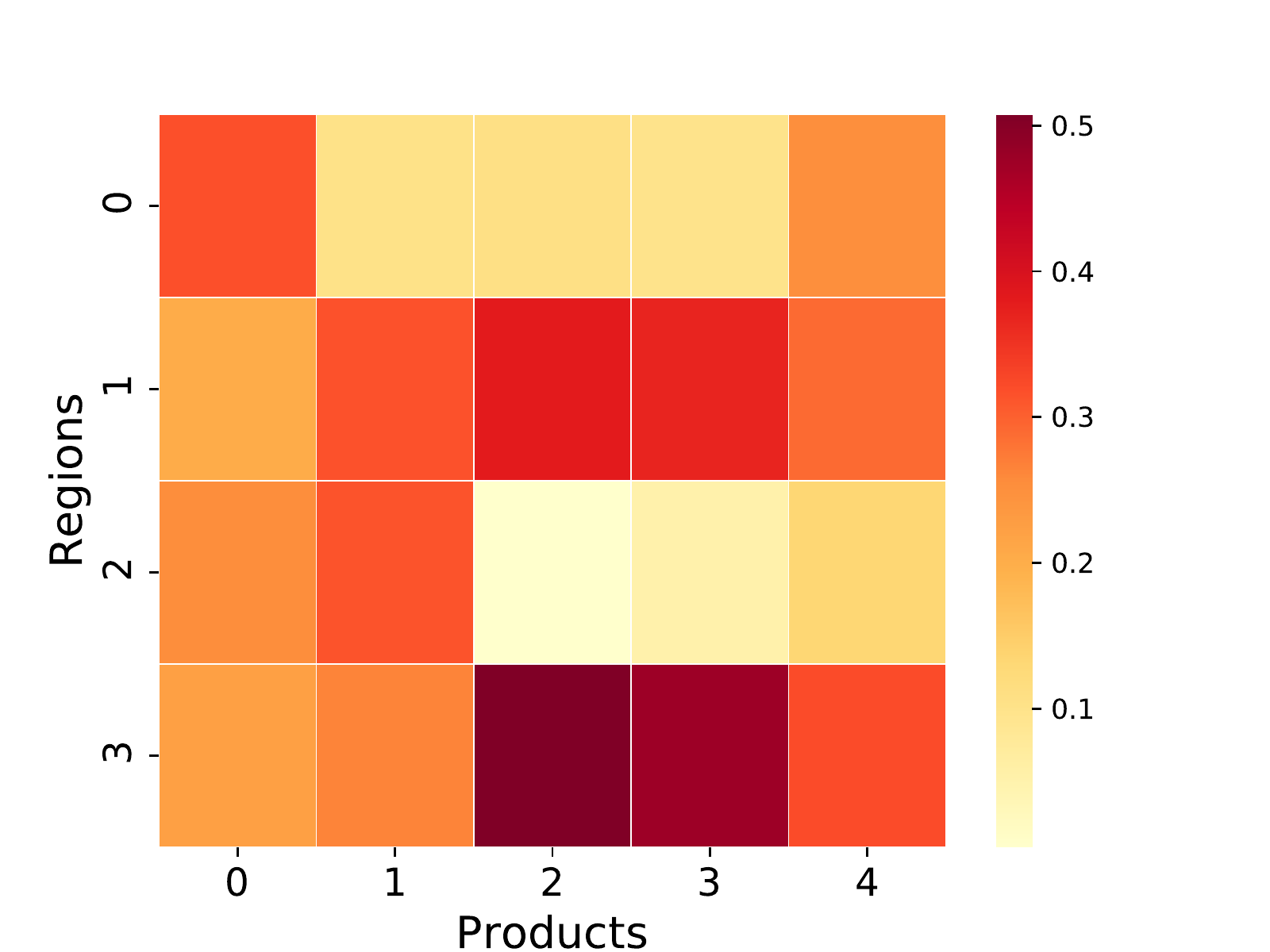}}

\caption{An example of the product allocation problem in physical retail. We provide a sample floor plan of a small, retail environment (a). Each section of the store is partitioned into ``regions''  (e.g., $r_1$). The product distributor or retailer has to choose the regions in which to put each of five possible products. The current product locations are plotted as colored $x$'s. We visualize the current allocation strategy as a state matrix, where blue components denote a given region, product combination has been selected (b). We also show the historical spatial distribution of revenue as a heat map (c). Darker colors indicate more historical revenue. The figure suggests that the current configuration may be sub-optimal. In reality, many large retail environments have thousands of products and many regions.}
\label{intro-fig}
\vspace{-4mm}
\end{figure}

Some existing work explores domains adjacent to the optimal product allocation problem. A large body of operations research analyzes shelf space distribution. For example, early work proposed a dynamic programming algorithm to discover an optimal shelf allocation strategy \cite{zufryden}. Other work poses shelf space allocation as a constrained optimization problem that can be solved via simulated annealing \cite{borin}. More contemporary studies propose frequent pattern mining approaches to determine profitable product item sets \cite{brijs} \cite{aloysius}. To the best of our knowledge, none of the existing literature has studied the spatial effects of product locations across the entire store.

However, learning a strategy for optimal product allocation is non-trivial. First, the number of candidate allocation strategies is large but the historical data usually only explores a small subset. Not to mention that sales are also correlated with other factors such as holidays and store promotions, which makes the search space even bigger. Because of this issue of data sparsity we cannot directly rely on historical data to learn the best strategy. Second, the cost of experimentation and exploration is high. It is not feasible to perform extensive experiments due to the potential lost revenue and the physical cost of moving products around the store. Finally, the correlation between product positions and sales is likely complex and non-linear due to the dynamic nature of the market; simple search heuristics may not provide an optimal policy. For all of these reasons, we need an approach that can accurately reflect the environment in a cost-efficient way.

Therefore, we design a new framework to solve these challenges. We propose a probabilistic spatial demand simulator to be a mirror of the real environment and act as a mechanism to study more complex search algorithms such as reinforcement learning without incurring the high cost of exploration in the physical world. We train the proposed model using a new, real-world dataset. Additionally, when deployed online, the model could be used to perform Monte Carlo rollouts for efficient exploration and experimentation \cite{kaiser}.

In our experiments, we demonstrate that the proposed model can effectively recover ground truth test data in two retail environments. Finally, we do a preliminary study into different optimization techniques using the proposed model.

In summary the key contributions of our paper are:

\begin{itemize}
    \item We study the new problem of optimal product allocation in physical retail
    \item We propose a probabilistic model of spatial demand that can accurately recover observed data, and generate data for new environment states
    \item We train PSD-sim on real data from two different retail stores
    \item We do a preliminary study into various optimization methods and show that Deep Q-Learning can learn an optimal allocation policy
\end{itemize}

\section{Problem Definition}\label{prob-def}
In the following section, we provide a formal definition of the optimal allocation problem. Additionally, we define the necessary components of our reinforcement learning agent: the state space, action space, reward function, and state transition function.
\subsection{Optimal Allocation Problem}

In a physical retail environment $\mathcal{R}$ with a set of $n$ spatial regions, we represent the environment with a spatial graph $\mathcal{R} = (\mathcal{V}, \mathcal{E})$, where each region $r_i\in \mathcal{V}$ is a vertex in the graph, the spatial neighboring relation between two regions $r_i$ and $r_j$ are represented as $e_{ij}\in \mathcal{V}$. From $\mathcal{G}$, we can construct the adjacency matrix, $\textbf{A}$.

Additionally, we observe a set of $k$ products, $\mathcal{M} = \{m_j : 0 < j <=k\}$ that are sold. For each product, $m_j$, we know the retail price, $p_j$. 

The decision process faced by the retailer is to allocate each product in $\mathcal{M}$ across regions in $\mathcal{R}$. We define the allocation policy as a function $f$:

\begin{equation}
    f: \mathcal{R} \times \mathcal{M} \rightarrow \mathcal{Z}
\end{equation}
\begin{equation}
    \mathcal{Z} = \{\langle r_i, p_j \rangle , ... \langle r_w, p_q \rangle \}
\end{equation}

Where $\mathcal{Z}$ is the set of selected product region, such that $w <= n$, $q <= k$ and $\mathcal{Z} \subseteq \mathcal{R} \times \mathcal{M}$. This function is typically dynamic over time, which we denote as $f^{t}$. To simplify computation, we treat $\mathcal{Z}^{t}$ as an $(n \times k)$ grid and refer to it as the board configuration at time, $t$. An optimal retail strategy is to find the allocation policy that maximizes revenue:

\begin{equation}
    f^{\ast} = \sum_{t}^{T} \argmax_{f^{t}} \sum_{i, j \in f^{t}(\mathcal{R}, \mathcal{M})} p_j q_i
\end{equation}

where $p_j$ is the price for product $m_j$, and $q_i$ is the quantity sold in region $r_i$ and $T$ is the future time horizon of analysis. The main idea of the current work is to discover the long-term, optimal allocation policy, $f^{\ast}$ from data.

\subsection{Optimal Allocation as a Markov Decision Process}
We believe that the optimal allocation problem is well suited for reinforcement learning because the RL agent is designed for sequential decision making that maximizes expected discounted reward over time. We frame the inputs as a Markov Decision Process (MDP). An MDP is defined by the tuple $\langle \mathcal{S}, \mathcal{A}, P, r, \delta  \rangle$, where $\mathcal{S}$ is the state space, $\mathcal{A}$ is the set of possible actions, $P$ is the (typically unkown) state transition function, $r$ is the reward function and $\delta \in [0,1]$ is the discount factor. 

\begin{itemize}
    \item \textbf{State} At each time, $t$, we observe the state of the retail environment, $\mathcal{E}$. We define the state, $s_t \in \mathcal{S}$, as the tuple of state features, $s_t = \langle \mathcal{Z}^{{t}}, d^{t}, \textbf{g}^{(t-1)}  \rangle$, where $\mathcal{Z}^{{t}}$ is the current board configuration, $d^t$ is the current day of the week (e.g., Sunday $\rightarrow$ 0), and $\textbf{g}^{(t-1)}$ is a vector denoting the revenue at the previous time, $(p_j q_i)^{(t-1)} \forall z \in \mathcal{Z}^t$

    \item \textbf{Action} We define the action space  $\mathcal{A} = \mathcal{R} \times \mathcal{M} \times \{-1, 1\} \cup \{0\}$, indicating ``to place'', ``take way'' or ``do nothing'' for each product, $m_j$ in each region, $r_i$.
    \item \textbf{Reward} The reward function in this case is the total product revenue at time $t$, constrained by the monetary cost, $c$, of placing a set of products in each region:
    \begin{equation}
        r(t) = \sum_{i=1}^n \sum_{j=1}^k p_j q_{ij}^{t} - c \sum_{i=1}^n \mathbbm{1}_{\mathcal{Z}}(r_i)
    \end{equation}
    
    \item \textbf{State transition function}: The state transition, $P$ is defined as $p(s^{t+1} | s^t, a^t): \mathcal{S} \times \mathcal{A} \times \mathcal{S} \rightarrow [0,1]$, which gives the probability of moving to state, $s^{(t+1)}$ given the current state and action. In the optimal allocation problem the exact transition function, $P$ is unknown since the current state, $s^t$ depends on the results of the previous time, $\textbf{g}^{(t-1)}$. We model this transition as a stochastic process.
\end{itemize}
\section{Proposed Method}\label{method}

\begin{figure}
\centering
\begin{tabular}{cc}
      \includegraphics[scale=.35]{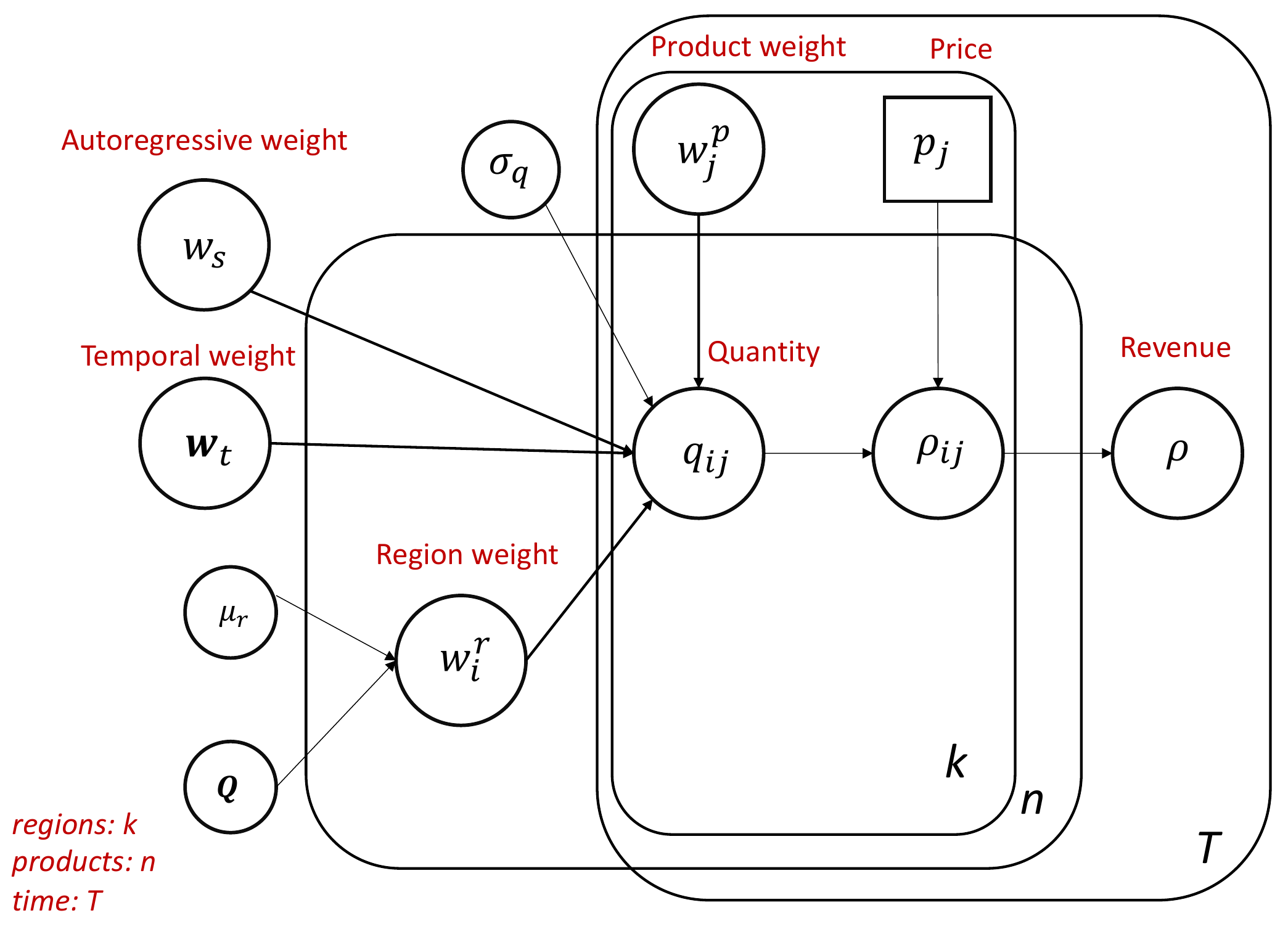}
\end{tabular}

\caption{An overview of the proposed model as a Bayesian network. The boxes are ``plates'' representing structures in the data. The plates marked by $k$, $n$ and $T$ represent products, regions, and time, respectively.  Circles denote random variables and squares are deterministic quantities. The model decomposes quantity as a function of region, product, time, and auto-regressive weights.}
\label{model-fig}
\end{figure}

In this section, we define our framework for solving the optimal allocation problem. Specifically, we outline our proposed environment model that is used to simulate spatial demand.

\subsection{Stochastic Model of Spatial Demand}\label{simple-model}

We propose the following stochastic model of spatial demand in physical retail. See Figure \ref{model-fig} for an overview. In the current work, the stochastic model is used as a `simulator' to enable offline policy learning. There are many advantages of using a probabilistic model in the optimal product allocation problem. First, we are able to incorporate prior knowledge about the data generating process, which can again improve data efficiency and model effectiveness. Second, it provides a natural framework for simulating future scenarios through Monte Carlo roll-outs.

Our ultimate objective is to maximize total revenue at time, $\rho^{(t)}$, which is defined as:

\begin{equation}
    \rho^{(t)} =  \sum_{i=1}^n \rho_i^{(t)}
\end{equation}

where $\rho_i^{(t)}$ is the revenue for region, $r_i$. Region-level revenue is calculated over products, $m_j$:

\begin{equation}
    \rho_i^{(t)} = \sum_{j=1}^k p_j q_{ij}^{(t)}
\end{equation}

The key variable of interest is, $q_{ij}^{(t)}$, the quantity sold for product, $m_j$, region, $r_i$, at time, $t$. We model  $q_{ij}^{(t)}$ as a truncated normal random variable:

\begin{equation}
    q_{ij}^{(t)} \sim \psi(\mu, \sigma, a, b)
\end{equation}

where, $\psi(\mu, \sigma, a, b)$ is the pdf of the truncated normal distribution. The term, $\phi(z)$ is the standard normal pdf, and $\Phi(z)$ is its cumulative distribution function. See \cite{Burkardt} for more details. We set $a = 0$ and $b=+\infty$, which forces $\Phi(\mu, \sigma^2; b) = 1$ and constrains quantity, $ q_{ij}^{(t)} \in \mathbb{R}^{+}$. The prior for $ q_{ij}^{(t)}$ is characterized by the mean, $\mu_q$, which is a linear function of environment features, $\textbf{x}$ and learned weights, $\textbf{w}$, and the inverse gamma distribution for the variance, $\sigma_q$:

\begin{equation}
    \mu_q = \textbf{x}^{\top}\textbf{w} + b
\end{equation}

\begin{equation}
    \sigma_q \sim \text{IG}(\alpha_q, \beta_q)
\end{equation}

In our environment, we observe temporal features, $\textbf{x}_t$,  region features, $\textbf{x}_r$, product features, $\textbf{x}_p$, and autoregressive features, $\textbf{x}_s$: \textbf{x} = $[\textbf{x}_t, \textbf{x}_r, \textbf{x}_p, \textbf{x}_s]^{\top}$. We discuss our feature extraction approach more in section \ref{features}

\subsubsection{Region-level Weights}
We initially model the weights for each spatial region with a multivariate normal distribution, with mean vector, $\boldsymbol{\mu}_r$ and covariance matrix, $\textbf{Q}_r$:

\begin{equation}\label{simple-w-r}
    \textbf{w}_r \sim \mathcal{N}(\boldsymbol{\mu}_r, \textbf{Q}_r)
\end{equation}

\subsubsection{Product-level Weights}

We also define weights for each product, $m_j$, as follows:

\begin{equation}
    \textbf{w}_p \sim \mathcal{N}(\boldsymbol{\mu}_p, \boldsymbol{\Sigma}_p)
\end{equation}

\begin{equation}
    \boldsymbol{\mu}_p \sim \mathcal{N}(\boldsymbol{\delta}_p, \boldsymbol{\Gamma}_p)
\end{equation}

\begin{equation}
    \boldsymbol{\Sigma}_p = \textbf{L}\textbf{L}^{\top} \sim \text{LKJ}(\sigma_p)
\end{equation}

We put a multivariate normal prior over the mean vector, $\boldsymbol{\mu}_p$ which has hyperparameters $\boldsymbol{\mu}_t$ and $\boldsymbol{\Sigma}_t$. Additionally, we put an LKJ prior over the covariance matrix, $\boldsymbol{\Sigma}_p$. We reparameterize $\boldsymbol{\Sigma}_t$ as its cholesky decomposition, $\textbf{L}\textbf{L}^{\top}$, so that the underlying correlation matrices follows an LKJ distribution \cite{lewandowski}. The standard deviations, $\sigma_p$, follow a half-cauchy distribution. The advantage of the LKJ prior is that is more computationally tractable than other covariance priors \cite{lewandowski}.

\subsubsection{Temporal weights}
 The temporal features capture the long-term and short-term seasonality of the environment. The temporal weights are defined similar to the product weights. Namely, the temporal weights, $\textbf{w}_t$, follow a multivariate normal distribution, with a normal prior over the mean, and the LKJ prior for the covariance matrix:

\begin{equation}
    \textbf{w}_t \sim \mathcal{N}(\boldsymbol{\mu}_t, \boldsymbol{\Sigma}_t)
\end{equation}

\begin{equation}
    \boldsymbol{\mu}_t \sim \mathcal{N}(\boldsymbol{\delta}_t, \boldsymbol{\Gamma}_t)
\end{equation}

\begin{equation}
    \boldsymbol{\Sigma}_t = \textbf{L}\textbf{L}^{\top} \sim \text{LKJ}(\sigma_t)
\end{equation}

\subsubsection{Autoregressive weight}
Finally, we specify the weight of previously observed revenue values on $q_{ij}^{(t)}$. The feature, $\textbf{x}_s$ is an autoregressive feature denoting the previous k values of product-level revenue, $\rho_j^{t} =  \sum_{j=i}^n p_j q_{ij}^{(t)}$. We assume truncated normal prior for $w_s$, and half cauchy priors for the location, $\mu_s$ and scale, $\sigma_s$:

\begin{equation}
    w_s \sim  \psi(\mu_s, \sigma_s, a, b)
\end{equation}

\begin{equation}
    \mu_s \sim \text{HalfCauchy}(\phi_s)
\end{equation}

\begin{equation}
    \sigma_s \sim \text{HalfCauchy}(\psi_s)
\end{equation}

We again set $a = 0$ and $b=+\infty$ such that $\textbf{w}_s \in \mathbb{R}^{+}$.

\begin{equation}\label{heterogenous-w-r}
    \textbf{w}_{ij}^r \sim \mathcal{N}(\textbf{w}_r, \textbf{Q}_r)
\end{equation}

\begin{equation}
    \textbf{w}_r \sim \mathcal{N}(\boldsymbol{\mu}_i, \textbf{Q}_r)
\end{equation}

Note that both $\textbf{w}_r$ and $\textbf{w}_{ij}^r$ share the same same covariance structure. Thus, the region weights are only hierarchical in their means. Additionally, we treat the upper-level mean vector, $\boldsymbol{\mu}_r$ as hyperparameter. In Section \ref{experiments} we test which environment model is more effective at predicting revenue on a test set.

\subsection{Training} We train the proposed model using the No U-Turn Sampler (NUTS) algorithm \cite{nuts}. This allows us to draw samples from the posterior distribution of model weights, $\textbf{W}$, as well as the posterior predictive distribution of quantity, $q_{ij}^{(t)}$, and revenue $\rho^{(t)}$. We use Automatic Differention Variational Inference (ADVI) \cite{advi} as an initialization point for the sampling procedure. All models are implemented in PyMC3 \cite{pymc3}

We initialize with ADVI using 200,000 iterations. Once initialized, we sample the posterior using NUTS with a tuning period of 5,000 draws followed by 5,000 samples across four chains.

\subsection{Feature Extraction}\label{features}
In order to train the proposed model, we extract environment-level features, $\textbf{x}$, which is composed of temporal features, $\textbf{x}_t$,  region features, $\textbf{x}_r$, product features, $\textbf{x}_p$, previous sales features and $\textbf{x}_s$.

\begin{itemize}
    \item \textbf{Temporal features} We use a one-hot vector denoting the day of the week for, $\textbf{x}_t$. This feature vector captures the short-term temporality commmon in physical retail settings. For example, weekends tend to be busier shopping days than weekdays. 
    \item \textbf{Region features} We again use a one-hot vector for spatial regions, $\textbf{x}_r$. This feature vector 'turns on' the weight that each region has on quantity via the weight vector, $\textbf{w}_r$.
    \item \textbf{Product features} We expect each product to vary in popularity. We capture this effect by constructing a one-hot vector for products, $\textbf{x}_p$.
    \item \textbf{Previous sales features} Finally, we construct an autoregressive sales feature that represents the sales at time, $t-1$. We use the previous sales for product $m_j$, summed across all regions, $w_s = \rho_j^{(t-1)} = \sum_{i=1}^k p_j q_{ij}^{(t-1)}$. This feature captures micro-fluctuations in demand for each product. 
\end{itemize}

\section{Experiments}\label{experiments}

In the following section we first describe the dataset and discuss interesting features of the problem. Next, we perform empirical evaluations of the proposed model across two large retail environment by showing that it can more accurately recover test data better than more elementary baselines.  We explore the model by discussing the estimation of region weights, and show that it is robust to previously unseen states. Finally, we do a preliminary inquiry into effective methods for  optimization.

\subsection{Dataset Decription}

\begin{table}[h]
    \centering
    \begin{tabular}{|c|c|c|c|}
         Store Id & Regions & Products & Time Horizon \\
         \hline
         \#1 & 17 & 15 & 8/2018 - 8/2019 \\
         \#2 & 12 & 15 & 8/2018 - 8/2019  \\
         \hline
    \end{tabular}
    \caption{Dataset Summary}
    \label{data}
\end{table}

\textbf{Stores}: We collect data from two large supermarket and retail stores in Salt Lake City, UT, USA. Each store primarily sells groceries, common household goods and clothing. Our dataset is comprised of transactions from August 2018 to August 2019.  \\
\textbf{Products}: We observe quantities sold for a set of 15 products, as well as each product's average price over the year. All of the products in our dataset are popular beverage products.

\textbf{Regions:} The data provides daily counts of quantities at the region-product level. Additionally, the locations of the products are varied in product "displays". These displays are small groups of products intended to catch the eye of the shopper. See Figure \ref{intro-fig} for an example of a product display layout. Store 1 is comprised 17 regions, and store 2 has 12. Each region represents a section of the store. In general regions tend to be constructed based the functionally of each space (e.g., pharmacy, deli, etc.). We construct a spatial graph of these regions.

\begin{figure}
    \centering
    \includegraphics[scale=.3]{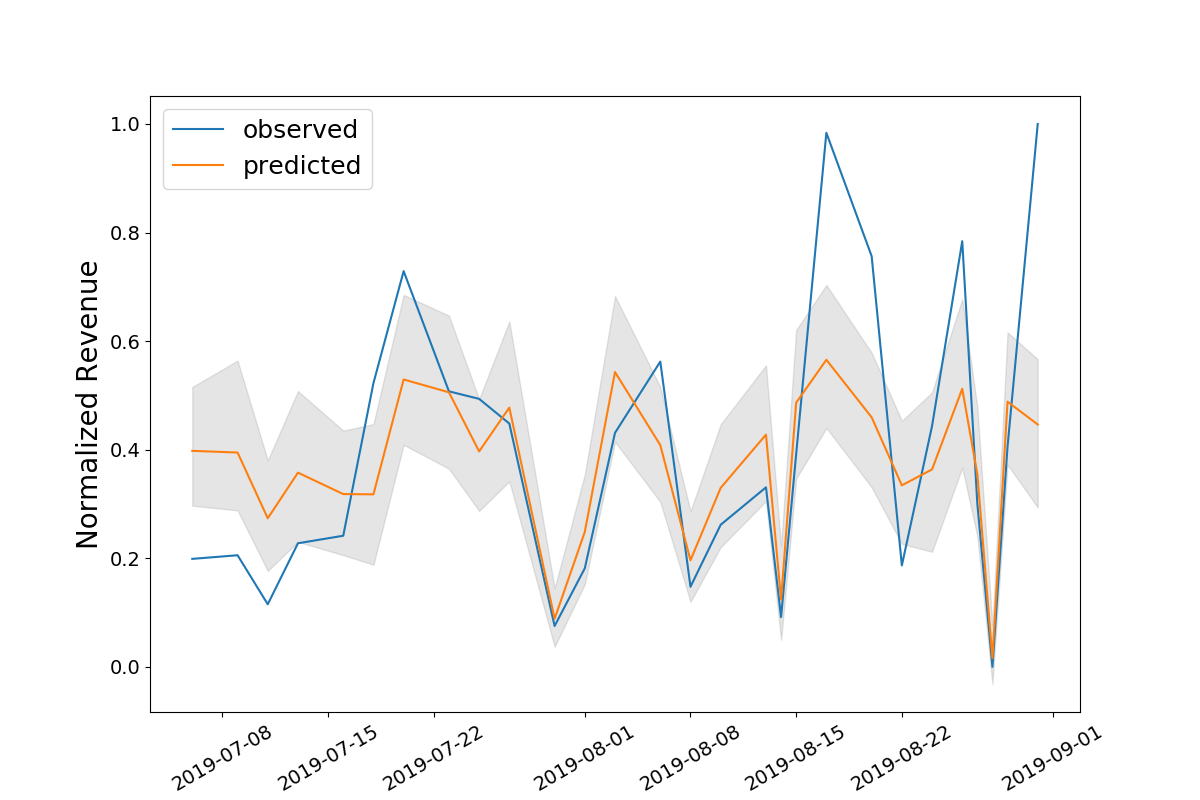}
    \caption{Predictions and observed revenue during the test period. Revenue is aggregated to the store-level. We display the results from store 2 above. We show the posterior distribution for revenue by plotting the mean (blue line) and inner 95\% credible interval (gray shaded area). In general, the predicted revenue mirrors the behavior of the ground truth data. the proposed model correctly predicts directional changes (i.e., positive or negative) 82\% of the time.}
\label{data-overview}
\end{figure}

\subsection{Model Evaluation}

We first evaluate the effectiveness of the proposed model in predicting revenue on a test dataset. Specifically, we partition the time series into a training period from August 1, 2018 - July 4, 2019 , and a test period of July 5, 2019 to August 31, 2019. We compare the proposed model to a variety of discriminitive baselines, and simpler variants of the proposed model. We evaluate all models in terms of the following error metrics:
\begin{equation}
    \text{MSE} =  \frac{1}{nkT} \sum_{t=1}^T\sum_{i=1}^n \sum_{j=1}^k (\rho^{(t)}_{ij} - \hat{\rho}^{(t)}_{ij})^2
\end{equation}

\begin{equation}
    \text{MAE} =  \frac{1}{nkT} \sum_{t=1}^T\sum_{i=1}^n \sum_{i=j}^k | \rho^{(t)}_{ij} - \hat{\rho}^{(t)}_{ij} |
\end{equation}

where the predicted revenue is equal to the quantity times price for the $i^{th}$ product, in the $j^{th}$ region, at time, $t$: $\hat{\rho}^{(t)}_{ij} = \hat{q}^{(t)}_{ij}p_j$. To compare to the discriminitive models, we obtain a point estimate for $\hat{q}^{(t)}_{ij}$ by computing the mean of the samples taken from posterior predictive distribution.

\begin{table}
        \centering
        \caption{Evaluation of the proposed model}
        \label{ppc-results}
        \begin{tabular}{c|c|c|c|c|}
             \cline{2-5}
              & \multicolumn{2}{c}{Store 1} & \multicolumn{2}{c|}{Store 2} \\
              Environment Model & MSE & MAE & MSE & MAE \\
            \hline
             OLS  & 2845.61 & 28.01 & 4816.41  & 34.81 \\
             RF  & 2908.73 & \textbf{26.77} & 5090.11  & 36.34 \\
             MLP  & 4037.91 &  34.66  & 7322.86  & 44.37 \\
             \hline
             Proposed  & \textbf{2615.32} & 27.67 & \textbf{4492.52}  & \textbf{34.48} \\
        \hline
        \end{tabular}
\end{table}

\subsubsection{Baseline Approaches}
The proposed model is a generative environment model and is able to draw samples from the full posterior distribution of revenue, $\rho^{(t)}$.  We also compare to the following discriminative prediction models:
\begin{itemize}
    \item \textbf{Linear Regression (OLS)}: Classical least squares regression that decomposes predicted quantity as a linear function of weights: $\hat{q}^{(t)}_{ij} = \textbf{X} \textbf{w} + b$.
    \item \textbf{Random Forest (RF)}: An ensemble regressor that learns many decisions trees and averages over the labels in each terminal node to compute, $\hat{q}^{(t)}_{ij}$. We use 100 trees.
    \item \textbf{Multilayer Perceptron (MLP)}: A simple neural network with two hidden layers of dimensions 256, and 128 with ReLU activations, MSE loss, and stochastic gradient descent optimizer.
\end{itemize}

We use the same features for all baselines. The features used in the experiment are described above.

\subsubsection{Results}

We report the results in Table \ref{ppc-results}. Additionally, predictions over the test set are plotted in Figure \ref{data-overview}. Overall we have the following observations from the experiment.

First, the proposed model is overall more accurate at predicting future states than baselines.  In particular, the proposed model yields the smallest MSE scores. MSE give a higher penalty to large errors, so in general the proposed model tends to make fewer, bad mistakes than all other baselines. This result holds both in store 1, and store 2. Additionally the proposed model minimizes the MAE score in store 2, but  is beat out by only the Random Forest baseline for store 1. Upon closer analysis we see that the Random Forest baseline has the second largest MSE score in store 1, which indicates that the Random Forest regressor has a higher variance than the proposed model. Overall, the proposed model is better or comparable to all baselines in both retail stores.

Second, the use of prior information in the proposed model allows it to perform better than the discriminitive baselines. Because the proposed model is a generative, Bayesian regression model we are able to set key hyperparameters at values according to our prior knowledge. For example, we know that retail sales increase on the weekends. By guiding the estimation of model parameters through the use of human knowledge the proposed is able to achieve prediction performance superior to OLS, RF, and the MLP in nearly all cases.

\begin{figure*}[t!]
\centering
\subfloat[Store 1]{\includegraphics[width=0.45\linewidth]{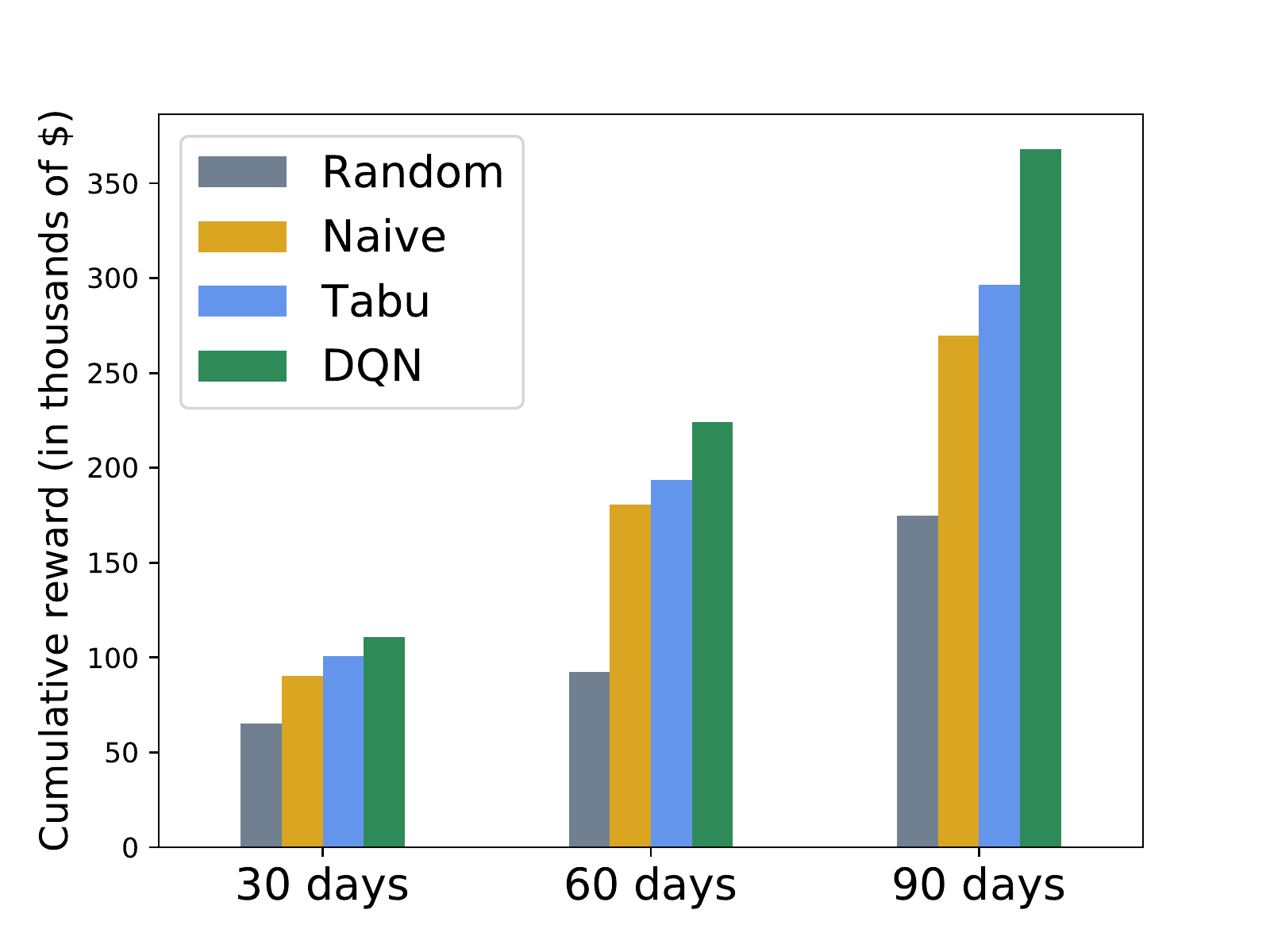}}
\subfloat[Store 2]{\includegraphics[width=0.45\linewidth]{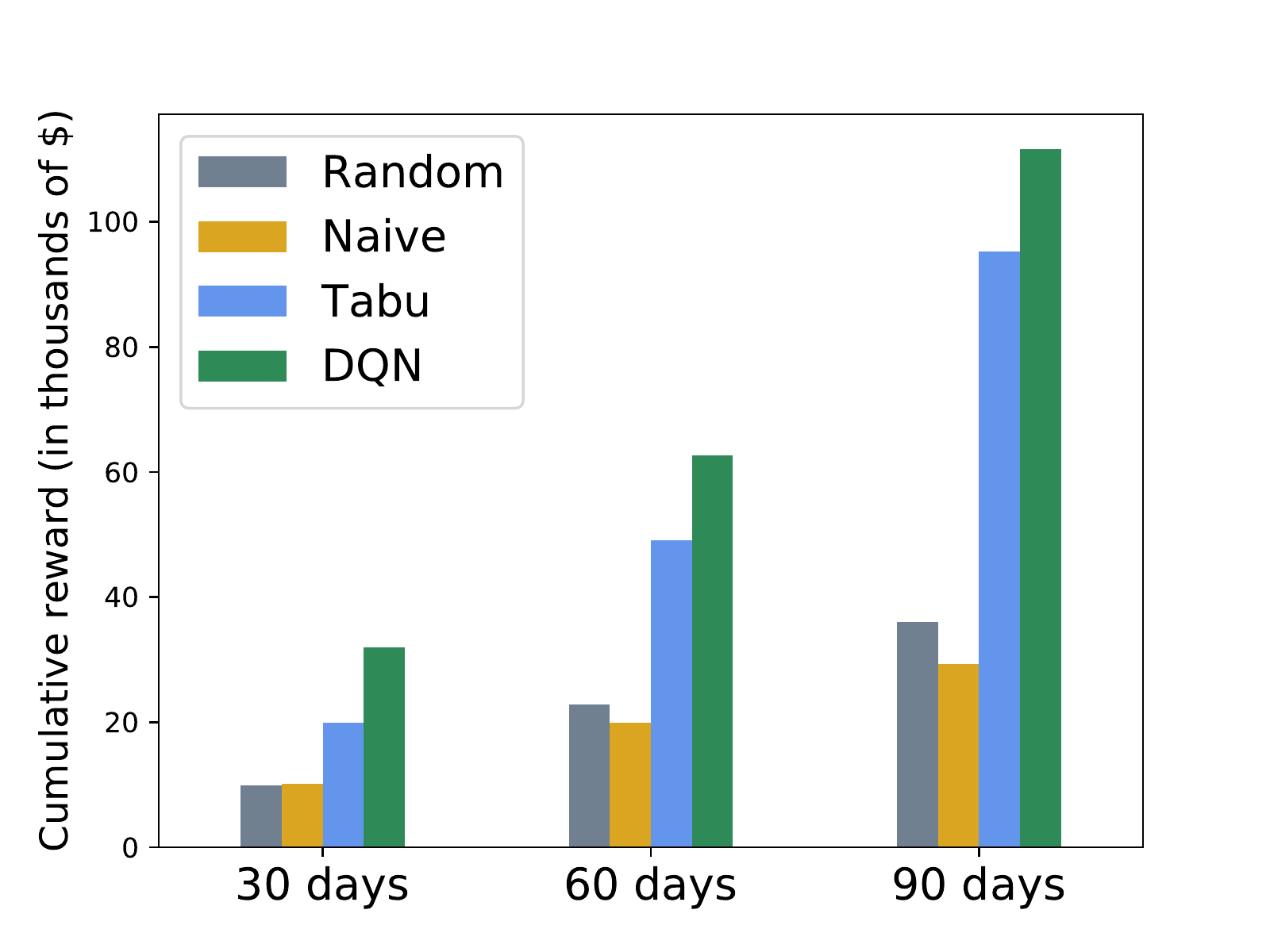}}

\vspace{-4mm}
\caption{A comparison of three search algorithms across store 1 and store 2. We vary the episode length in 30 day increments (i.e., 30, 60, and 90 days in the future). The DQN algorithm is superior in all cases. Additionally, we observe that as the episode length increases so does the relative effectiveness of the DQN. The DQN agent excels in the longer episode settings because it is able to learn important, longer term strategies. On average, DQN offers an improvement of 24.5\% over Tabu search in terms of cumulative test reward.}
\label{optim}
\vspace{-4mm}
\end{figure*}

\subsection{Optimization Techniques}

In this section we perform a preliminary study into various search algorithms to solve the optimal product allocation problem with the the proposed model environment model. Because exploration and experimentation in the physical world is costly, it is often preferable to design an agent that can learn a policy offline before deploying into the online environment \cite{kaiser}.

\subsubsection{Search Algorithms}
To this end we compare four methods to search the problem  space: random search, naive search, Tabu search, and Deep Q-Learning
\begin{itemize}
    \item \textbf{Random Search} A search algorithm that relies on a totally random policy: at each time step, $t$ choose a random action.
    \item \textbf{Naive Search} The naive strategy in this case is simply ``do nothing." At each time step, we do not move any products and do not deviate from the initialized allocation policy. This baseline allows us to assess whether searching and exploration is useful at all.
    \item \textbf{Tabu Search}: A local neighborhood search algorithm that maintains a memory structure called a ``Tabu'' list. The ``Tabu'' list is comprised of recent actions to encourage exploration and avoid getting trapped in local maxima. We implement the Tabu algorithm with a ``Tabu'' list of the previous 50 actions. We treat the local neighborhood search as the enumeration over set of feasible actions given the current state, $s_t$.
    \item \textbf{Deep Q-Learning (DQN)}: A reinforcement learning algorithm that utilizes a neural network to approximate the state-action function, $Q(s, a)$. The DQN typically employs an $\epsilon\text{-greedy}$  strategy for exploration. The exploration probability, $\epsilon$ is typically annealed throughout training. DQN has been shown to be effective for learning policies in complex, dynamic environments such as Atari \cite{mnih}, Go \cite{silver-16} \cite{silver-17}, and ride dispatching \cite{lin-msu}, and traffic signal control \cite{intellilight}. We train our DQN using 50,000 training iterations prior to the test period.
\end{itemize}

\subsubsection{Policy Evaluation}

In this section we conduct a policy evaluation experiment. We randomly fix the initial environment state and allow each of the search algorithms listed above to interact with the environment according to its corresponding strategy in a test period of one episode. The state in store 1 is initialized with 96 product-region pairs, while the state in store 2 has 30. We record the total reward accumulated by each agent during the entire episode. For each store, we vary the episode length in 30 day increments: 30, 60, and 90 days in the future. This allows us to evaluate whether longer rollouts have an effect on the policy of each agent. The results of the policy evaluation experiment are reported in table \ref{optim}.

In general, we see that DQN is the most effective search algorithm in both stores, and across all three episode settings. In each case, it accumulates the most total reward in the test episode. On average, DQN is 24.5\% better than Tabu, in terms of cumulative test reward. Tabu is the second most effective search strategy, beating out the random and naive search heuristics in all cases. Interestingly, the naive search baseline of ``do nothing'' is more effective than random searching in store 1, but not in store 2. 

Additionally, it appears that as the episode length is increases, so too does the relative effectiveness of DQN as compared to Tabu. In the store 1, 30 day episode setting, DQN exceeds Tabu by \$10k. This difference increases to \$30k for 60 days and \$72k for 90 days. In store 2 we see a similar effect. The difference between DQN and Tabu increases from \$12k to \$13.5k to \$16k in the 30, 60, and 90 day settings respectively. Not only is DQN more effective, but its performance relative to other baselines gets better with longer episodes. 

DQN excels as episode length increases in large part because the underlying $Q$-function is an approximation of discounted, expected reward over time. This allows the agent to potentially think multiple steps ahead and take a set of actions that yield low immediate reward, but higher reward in later steps. Conversely, the random and Tabu search baselines are short-term or greedy search algorithms. Especially in the case of Tabu; at each time step, an action is solely selected based on what will maximize short-term reward. These results suggest that the correlations between spatial allocation and sales is complex and dynamic. Thus both of the two baselines achieve sub-optimal policies.

It is also interesting to note the behavior of the naive search compared to the random strategies across the two stores. In store 1, the environment is initialized with an allocation strategy that already has many product placements (96). We see that the naive strategy is a strong baseline, and is superior to the random policy in each of the 30, 60 and 90 day settings. However, in store 2 where the initial allocation is more sparse (30 placements), the random policy is better than or equal to the naive search. This suggest that as more products are placed it is more difficult to find incremental improvements in the allocation strategy.

\section{Related Work}

There are two major streams of literature that intersect with our problem: 1) shelf space allocation and 2) deep reinforcement learning for spatial allocation.

\subsection{Shelf Space Allocation} 
The shelf space allocation allocation problem has been studied in the operations research literature for many decades. Some classical work approaches the problem by proposing a dynamic programming algorithm to allocate limited shelf space among a finite set of products. In this case, the objective function is composed of revenue, costs and a set of constraints \cite{zufryden}. Later work proposed a simulated annealing optimization approach that accounts for two primary decisions variables: product assortment and allocated space for each product \cite{borin}. This optimization technique accounts for many different environment variables such as item profitability, brand elasticities, and supply chain features. More recently, frequent pattern mining algorithms have been proposed to allocate product shelf space. For instance Brijs et al. \cite{brijs} propose the PROFSET algorithm, which an association rule algorithm that mines customer basket sets to identify profitable product pairings. This algorithm is a extension of frequent item set algorithms that also accounts for product value. Extensions of this idea have also been proposed. Aloysius and Binu propose a PrefixSpan algorithm for shelf allocation  that first identifies complementary categories from historical purchase data before identifying product mix strategies within categories \cite{aloysius}.

These existing studies differ from our work in the following ways. First, they all focus on micro-regions (shelves) within the retail environment. The spatial effects these models capture are markedly different from the macro-level ones tackled in the current work. Second, these studies focus on the number of each product on a shelf. They try to maximize profitability given the fixed shelf volume. This optimization problem is fundamentally different from allocating products across the entire store. For these reasons, none of these methods can be directly applied to our problem.

\subsection{Deep Reinforcement Learning for Spatial Resource Allocation} Recent breakthroughs in reinforcement learning \cite{mnih} \cite{silver-16} \cite{silver-17} have spurred interest in RL as an optimization approach in complex and dynamic environments. In particular, recent studies have proposed RL algorithms as a mechanism for  spatiotemporal resource allocation.

\textbf{Order dispatching.} Significant attention has been paid to the order dispatching problem in ride sharing systems. Briefly, order dispatching refers to the problem of efficiently matching riders and drivers in an urban environment. The RL agent must learn the complex spatial dynamics to learn a policy to solve the dispatching problem. For example, Lin et al. \cite{lin-msu} tackle the dispatch problem by proposing a contextual multi-agent reinforcement learning framework that coordinates strategies among a large number of agents to improve driver allocation in physical space. Additionally, Li et al. \cite{li} also approach the order dispatching problem with multi-agent reinforcement learning (MARL). Their method relies on the mean field approximation to capture the dynamic, spatially distributed fluctuations in supply and demand. They empirically show that MARL can reduce supply-demand gaps in peak hours.

\textbf{Traffic signal control} Increasing traffic congestion is a key concern in many urban areas. Recent efforts to optimize traffic control systems via reinforcement learning has shown encouraging results. These systems seek to adjust traffic lights to real-time fluctuations in traffic volume and road demand. Wei et al \cite{intellilight} propose IntelliLight, which is a phase-gated deep neural network that approximates state-action values. More recently \cite{colight} proposes a graph attentional network to facilitate cooperation between many traffic signals.

\textbf{Spatial demand for electronic tolls} Chen et al. \cite{chen} propose a dynamic electronic toll collection system that adjusts to traffic patterns and spatial demand for roads in real time. Their proposed algorithm, PG-$\beta$, is an extension of policy gradient methods and decreases traffic volume and travel time.

While these reinforcement learning methods deal with the large-scale optimization of spatial resource, they cannot be directly applied to the product allocation problem because the all rely on domain-specific simulators. We propose our model in an effort to extend these state-of-the-art optimization techniques to our problem.
\section{Conclusion}

In this paper, we studied the automation of product placement in retail settings. The problem is motivated by the fact that well placed products can maximize impulse buys and minimize search costs for consumers. Solving this allocation problem is difficult because location-based, historical data is limited in most retail settings. Consequently, the number of possible allocation strategies is massive compared to the number of strategies typically explored in historical data. Additionally, it is generally costly to experiment and explore new policies because of the economic costs of sub optimal strategies, and operational cost of deploying a new allocation strategy. Therefore, we propose a probabilistic environment model called that is designed to mirror the real world, and allow for automated search, simulation and exploration of new product allocation strategies. We train the proposed model on real data collected from two large retail environments. We show that the proposed model can make accurate predictions on test data. Additionally, we do a preliminary study into various optimization methods using the proposed model as a simulator. We discover that Deep $Q$-learning techniques can learn a more effective policy than baselines. On average, DQN offers an improvement of 24.5\% over Tabu search in terms of cumulative test reward.

\bibliographystyle{aaai}
\bibliography{main}

\end{document}